# Detecting low-complexity unobserved causes

Dominik Janzing[1], Eleni Sgouritsa[1], Oliver Stegle[1,2], Jonas Peters[1], Bernhard Schölkopf[1]

[1] Max Planck Institute for Intelligent Systems, Tübingen, Germany
[2] Max Planck Institute for Developmental Biology, Tübingen, Germany
{janzing, sgouritsa, stegle, peters, bs}@tuebingen.mpg.de

## Abstract

We describe a method that infers whether statistical dependences between two observed variables $X$ and $Y$ are due to a "direct" causal link or only due to a connecting causal path that contains an unobserved variable of low complexity, e.g., a binary variable.
This problem is motivated by statistical genetics. Given a genetic marker that is correlated with a phenotype of interest, we want to detect whether this marker is causal or it only correlates with a causal one. Our method is based on the analysis of the location of the conditional distributions $P(Y|x)$ in the simplex of all distributions of $Y$. We report encouraging results on semi-empirical data.

## 1 Introduction

Statistical dependences between two variables $X$ and $Y$ indicate that (A) $X$ causes $Y$, (B) $Y$ causes $X$, or (C) there is a third variable $W$ influencing both $X$ and $Y$ [Reichenbach, 1956]. The case where dependences are generated by selection bias via implicit conditioning on a common effect of $X$ and $Y$ is excluded throughout the paper. In many applications, the time order or other prior knowledge excludes case (B). The distinction between cases (A) and (C) is a challenging task of causal inference. Note, however, that it is unsolvable if $W$ and $X$ are so strongly coupled that they attain identical values. $W$ could be, for instance, a physical quantity like temperature and $X$ be the value of $W$ shown by the measurement instrument. Every method therefore needs a sufficiently non-trivial relation between $W$ and $X$ (e.g. measurement error) leaving some sort of "fingerprint" on the distribution $P(X,Y)$.

After observing variables other than $X$ and $Y$, the problem of distinguishing between (A) and (C) can be addressed via conditional independences [Pearl, 2000] even if $W$ is unobserved. Detecting latent common causes (confounders) if only $X$ and $Y$ are observed requires strong assumptions on the data generating process. Known methods include linear relations between real-valued non-Gaussian variables (e.g. [Shimizu et al., 2009]) and non-linear relations with additive noise [Janzing et al.].

We describe a method that uses $P(X,Y)$ to infer whether an unobserved "low complexity" variable $Z$ is contained in all unblocked causal paths between $X$ and $Y$ or whether the link is "direct" in the sense that no such variable exists. Here, low complexity is typically understood in the sense of low range (e.g. a binary variable), but the definition can also include variables with compact range, which can be detected under certain conditions. This inference problem is quite different from the problem of distinguishing between (A) and (C), because $Z$ could occur, for instance, as $X \to Z \to Y$ (corresponding to (A)) or as $X \leftarrow Z \to Y$ (corresponding to (C)). However, below we describe an application from statistical genetics where a "direct" link provides evidence for (A), whereas a low range variable contained in the causal paths is taken as a hint for (C). This is because, in this application, we consider low range variables more likely to be on confounding paths than mediating the influence of $X$ on $Y$.

An important problem in biology and medicine is to find genetic causes of phenotypic differences among individuals. Let $Y$ describe a phenotypic difference among individuals such as the presence or absence of a disease, the size of a plant, or the expression level of a gene. These phenotypes are known to correlate with polymorphic loci in the genome, such as single-nucleotide polymorphisms (SNPs). However, due to the strong dependences among nearby SNPs, it is hard to identify those that influence the phenotype. Depending on the genetic architecture, SNPs can be encoded in a number of ways. Here, we choose a binary encoding, with 0 corresponding to the "common" genetic configuration and 1 corresponding to the "less frequent" variant. If $X$ describes the SNP under consideration, the task is to decide whether the dependence between $X$ and $Y$ is because $X$ influences $Y$ or only due to statistical dependence between $X$ and some unobserved SNPs $Z$ influencing $Y$ ($Z$ is low range if it describes only a small number of SNPs, each of which is a binary variable). $Z$ could be also some environmental condition that influenced $X$ (via evolution) and $Y$, provided that $Z$ has low range. Thus we have either $X \to Y$ or $X \leftrightarrow Z \to Y$ or $X \leftarrow Z \to Y$, where $\leftrightarrow$ symbolizes that $X$ and $Z$ are related by a common cause.

The remainder of this manuscript is organized as follows. Section 2 introduces the mathematical property of conditionals that we argue to provide evidence for a "direct" causal connection. Section 3 describes how to obtain evidence against a "direct" connection and how to gain information about the low range latent variable $Z$. Section 4 sketches possible methods to estimate the mathematical properties from finite data. Finally, Section 5 presents some experiments and the paper is concluded with a discussion in Section 6.

## 2 Evidence for a "direct" causal relation: pure conditionals

We first introduce a property of $P(Y|X)$ that we consider as providing evidence for a "direct" causal link $X \to Y$. Let $\mathcal{X}, \mathcal{Y}, \mathcal{Z}$ denote the ranges of $X, Y, Z$, respectively and $\mathcal{P}_X, \mathcal{P}_Y, \mathcal{P}_Z$ denote the simplex of probability distributions on these sets, respectively. Clearly, $P(Y|x) \in \mathcal{P}_Y$ for every $x \in \mathcal{X}$ and also every *convex* combination of distributions $P(Y|x)$ lies in $\mathcal{P}_Y$. Whether also *affine* combinations that contain some negative coefficients yield distributions in $\mathcal{P}_Y$ is an interesting property of $P(Y|X)$. Throughout the paper, we assume that $P(Y|X)$ has a density $p(y|x)$ that is continuous in both $x$ and $y$. Note that discrete variables are also included because the probability mass function (which is the density with respect to the counting measure) is always continuous with respect to the standard discrete topology.

**Definition 1 (pure conditionals)**
*A conditional $P(Y|X)$ is called $k$-wise pure if for every $k$-tuple of different $x$-values $(x_1, \ldots, x_k)$ the following condition holds: for all $\lambda \in \mathbb{R}^k \setminus [0,1]^k$ with $\sum_j \lambda_j = 1$*

$$\exists y: \quad \sum_{j=1}^k p(y|x_j)\lambda_j < 0$$

*We also say "pairwise pure" instead of "2-wise pure". $P(Y|X)$ is called one-sided pairwise pure if for every*

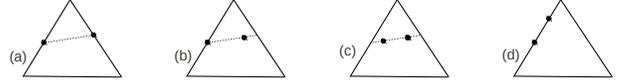

Figure 1: Visualization of the location of different $P(Y|x)$ in the simplex $\mathcal{P}_Y$, here for $|\mathcal{Y}| = 3$: (a) pairwise pure, because the line connecting $P(Y|x_1)$ and $P(Y|x_2)$ (the black dots) cannot be extended without leaving the simplex; (b) one-sided pairwise pure; (c) and (d) are not pairwise pure, although both points in (d) are not in the interior of $\mathcal{P}_Y$.

*pair $(x_1, x_2)$ with $x_1 \neq x_2$ and for all $\mu < 0$ either*

$$\exists y: \quad \mu p(y|x_1) + (1-\mu)p(y|x_2) < 0$$
$$\text{or } \exists y: \quad \mu p(y|x_2) + (1-\mu)p(y|x_1) < 0$$

*(see Fig. 1 for some examples).*

The class of pairwise pure conditionals will play a crucial role henceforth:

**Hypothesis 1 (causal relevance and purity)**
*(i) Many, but not all, interesting causal mechanisms $X \to Y$ in nature generate pairwise pure conditionals $P(Y|X)$ provided that $Y$ has large or even continuous range.*

*(ii) Observing that a conditional $P(Y|X)$ is pairwise pure provides some evidence for a "direct" causal link $X \to Y$. The existence of a low range variable $Z$ such that $X \perp\!\!\!\perp Y | Z$ can be even excluded under mild assumptions.*

*(iii) $P(Y|X)$ can still be pairwise pure if, apart from the "direct" link $X \to Y$, there exist additional unblocked paths between $X$ and $Y$ that contain a low range variable.*

The remainder of this section is devoted to supporting this hypothesis by several theoretical results. We prove some conditions for purity.

**Lemma 1 (quotient of densities)**
*$P(Y|X)$ is pairwise pure if and only if for every pair $(x_1, x_2)$ with $x_1 \neq x_2$*

$$\inf_{y \in \mathcal{Y}\,:\, p(y|x_2) \neq 0} \frac{p(y|x_1)}{p(y|x_2)} = 0. \quad (1)$$

*One-sided pairwise purity holds if and only if, for every pair $(x, x')$, (1) holds either for $x_1 = x$ and $x_2 = x'$ or for $x_1 = x'$ and $x_2 = x$.*

Proof: If (1) does not hold we set $c := \inf_y p(y|x_1)/p(y|x_2)$ with $0 < c < 1$. Then choosing

the coefficient $\mu = 1/(1-c)$ (such that $1 - \mu$ is negative) ensures

$$\mu p(y|x_1) + (1-\mu) p(y|x_2) \geq 0, \quad (2)$$

for all $y$ with $p(y|x_2) \neq 0$. If $p(y|x_2) = 0$, the left hand side of (2) is non-negative anyway. Hence, purity is violated. On the other hand, if $P(Y|X)$ is not pure there is by definition a pair $(x_1, x_2)$ and $\mu < 0$ such that $(1-\mu) p(y|x_1) + \mu p(y|x_2) \geq 0$ for all $y$, then $\frac{p(y|x_1)}{p(y|x_2)} \geq \frac{-\mu}{1-\mu}$, which contradicts (1). $\square$

If $\mathcal{Y}$ is finite, pairwise purity can easily be characterized:

**Lemma 2 ($Y$ with finite range)**
Let $|\mathcal{Y}|$ have finite range, then $P(Y|X)$ is pairwise pure if and only if for all pairs $(x, x')$

$$\operatorname{supp} p(y|x) \not\subseteq \operatorname{supp} p(y|x'),$$

where $\operatorname{supp} p(y|x) := \{y \mid p(y|x) > 0\}$ denotes the support of $p(y|x)$.

Proof: If the support of $p(y|x)$ is contained in the support of $p(y|x')$ then $y \mapsto p(y|x')/p(y|x)$ attains a non-zero minimum and the statement follows from Lemma 1. $\square$

Motivated by this result, we define $m(k)$ to be the maximal number of non-empty subsets of $\{1, \ldots, k\}$ such that no subset is contained in another one (obvious lower bounds for $m$ are therefore given by binomial coefficients) and obtain:

**Corollary 1 (purity for finite $|\mathcal{X}|$ and $|\mathcal{Y}|$)**
There exist pairwise pure conditionals $P(Y|X)$ for $|\mathcal{X}| \leq m(|\mathcal{Y}|)$, but not for $|\mathcal{X}| > m(|\mathcal{Y}|)$.

Corollary 1 shows that purity of $P(Y|X)$ requires the range of $Y$ to be large enough.

**Lemma 3 ($Y$ with compact range)**
If $\mathcal{Y}$ is compact and all densities $\{p(y|x)\}_{x \in \mathcal{X}}$ are strictly positive then $P(Y|X)$ is not even one-sided pairwise pure.

Proof: The function $y \mapsto p(y|x)/p(y|x')$ is strictly positive and continuous and thus attains a non-zero minimum. $\square$

The following Lemmas (4 and 5) support statement (i) of Hypothesis 1. Lemma 4 shows that additive noise models (which have already been proposed as natural models for causal relations [Hoyer et al., 2009]), are pairwise pure under some additional conditions:

**Lemma 4 (purity of additive noise models)**
Let $X, Y, E$ be real-valued variables. Let $E$ have strictly positive differentiable density. Then the additive noise model (ANM)

$$Y = f(X) + E \text{ with } E \perp\!\!\!\perp X$$

defines a pairwise pure conditional provided that $f$ is injective and

$$\frac{d \log p(e)}{de} \to \pm\infty \text{ for } e \to \mp\infty. \quad (3)$$

If $d \log p(e)/de$ is bounded from below or from above, the ANM is not pairwise pure.

Proof: We show that the quotient $p(y|x)/p(y|x')$ tends to zero or infinity for $y \to \infty$ or $y \to -\infty$. Introducing $\psi(e) := \log p(e)$ we have

$$\log \frac{p(y|x)}{p(y|x')} = \log \frac{p(e = y - f(x))}{p(e = y - f(x'))}$$
$$= \frac{d}{de} \psi(\tilde{e})(f(x) - f(x')),$$

where $\tilde{e}$ is some point between $y - f(x)$ and $y - f(x')$ (by the mean value theorem). For $y \to \pm\infty$ the logarithm of the quotient thus converges to $\mp\infty$ or $\pm\infty$, depending on the sign of $f(x) - f(x')$. Likewise, if $\psi'$ is bounded, the quotient $p(y|x)/p(y|x')$ has a strictly positive lower bound. $\square$

**Lemma 5 (Gaussian conditionals)**
For every $x$, let $P(Y|x)$ be a Gaussian with mean $\mu_x$ and standard deviation $\sigma_x$. If $\sigma_x = \sigma$ for all $x$ and $\mu_x \neq \mu_{x'}$ for $x \neq x'$ then $P(Y|X)$ is pairwise pure. If $(\mu_x, \sigma_x) \neq (\mu_{x'}, \sigma_{x'})$ for $x \neq x'$, then $P(Y|X)$ is one-sided pairwise pure.

Proof: $p(y|x)/p(y|x')$ is proportional to

$$\exp\left(-\frac{y^2(\sigma_x^2 - \sigma_{x'}^2) - 2y(\mu_x \sigma_{x'}^2 - \mu_{x'} \sigma_x^2)}{2\sigma_x^2 \sigma_{x'}^2}\right). \quad (4)$$

We first consider the case $\sigma_x \neq \sigma_{x'}$. Then either (4) or its inverse goes to 0 either for $y \to \pm\infty$ depending on which variance is larger. In the first case, $\lambda p(y|x) + (1-\lambda) p(y|x')$ is not a probability density if $\lambda < 0$ and in the second case for $\lambda > 1$. If $\sigma_x = \sigma$ for all $x$, we can write $Y = f(X) + E$ with Gaussian noise $E$. By assumption, $f : \mathcal{X} \to \mathbb{R}$ is injective and pairwise purity follows from Lemma 4. $\square$

It should be emphasized that the question of whether a conditional $P(Y|X)$ is pure is completely different from asking whether there is a decomposition of each $P(Y|x)$ into mixture components like Gaussians. One can easily construct pairwise pure conditionals where each $P(Y|x)$ is multi-modal. As opposed to the number of mixture components, purity is invariant with respect to parameter transformations on $Y$:

**Lemma 6 (parameter transformations)**
Let $g$ be a continuous bijection of $\mathcal{Y}$, then $P(g(Y)|X)$ is $k$-wise pure if and only if $P(Y|X)$ is.

The proof is obvious. The result shows that the class of post-nonlinear models [Zhang and Hyvärinen]

$$Y = g(f(X) + E) \text{ with } E \perp\!\!\!\perp X,$$

where $g$ is a bijection, is pairwise pure if and only if the corresponding additive noise model $Y = f(X) + E$ is pairwise pure.

There are, however, interesting noise distributions that do not render the additive noise model pairwise pure: Let the density of $p(y|x)$ satisfy for every $x$ the asymptotical property

$$p(y|x) \propto |y|^{-(1+\alpha_x)} \text{ for } y \to \pm\infty, \quad (5)$$

for some $\alpha_x > 0$. Distributions of this kind are sometimes called *fat-tailed*.

**Lemma 7 (fat-tailed conditionals)**
(i) If there is a pair $(x, x')$ such that $p(y|x)$ and $p(y|x')$ satisfy (5) with the same rate, then $P(Y|X)$ is not pairwise pure.

(ii) If for all pairs $(x, x')$ such that $P(Y|x)$ and $P(Y|x')$ satisfy (5), the rates are different, then $P(Y|X)$ is one-sided pairwise pure.

Proof: We can apply Lemma 1 to obtain $p(y|x)/p(y|x') \propto |y|^{\alpha_{x'} - \alpha_x}$ for $y \to \infty$. □

Hence, non-pairwise-purity does not disprove that $X \to Y$. Our claim is a cautious version of the converse statement: we argue that pairwise purity provides some evidence for a "direct" link between $X$ and $Y$. To this end, we discuss under which conditions observing that a conditional is pairwise pure can exclude the existence of a low complexity variable $Z$ on the causal paths:

**Lemma 8 (concatenation)**
If $X \perp\!\!\!\perp Y | Z$ then $P(Y|X)$ can only be $k$-wise pure if $P(Z|X)$ is $k$-wise pure.

Proof: If $P(Z|X)$ is not $k$-wise pure, we can find partially negative coefficients $\lambda_j$ with $j = 1, \ldots, k$ for which $\sum_{j=1}^{k} P(Z|x_j) \lambda_j \in \mathcal{P}_Z$, which implies $\sum_{j=1}^{k} P(Y|x_j) \lambda_j = \sum_{j=1}^{k} \int P(Y|z) dP(z|x_j) \lambda_j \in \mathcal{P}_\mathcal{Y}$. □

Below, Theorems 1 and 2 support part (ii) of Hypothesis 1: observing that a conditional $P(Y|X)$ is pairwise pure provides some evidence for $X \to Y$.

**Theorem 1 (excluding finite $Z$)**
If $P(Y|X)$ is pairwise pure then there is no variable $Z$ with $m(|\mathcal{Z}|) < |\mathcal{X}|$ such that $X \perp\!\!\!\perp Y | Z$. In particular, if $|\mathcal{X}|$ is infinite there is no $Z$ with finite range such that $X \perp\!\!\!\perp Y | Z$.

The proof follows easily by combining Corollary 1 with Lemma 8. A nice implication of Theorem 1 for real-valued $X$ concerns the additive noise model defined in Lemma 4: since $P(Y|X)$ is pairwise pure, there is no variable $Z$ with finite range such that $X \perp\!\!\!\perp Y | Z$.

Recalling our motivating biological application, Theorem 1 shows also that, if $P(Y|X)$ is a pairwise pure conditional with $X$ consisting of two or more binary variables, then there is no binary $Z$ in the connecting path (since $m(2) = 2 < 2^2$).

However, even if the range of $Z$ is so large that Theorem 1 is void, the existence of $Z$ can be nevertheless excluded by pairwise purity under a weak assumption:

**Theorem 2 (compact $Z$)**
If $P(Y|X)$ is pairwise pure then there is no variable $Z$ with compact range and $P(Z|x)$ having continuous strictly positive densities (or probability mass functions if $\mathcal{Z}$ is finite) such that $X \perp\!\!\!\perp Y | Z$.

The proof is given by combining Lemma 3 with Lemma 8. If both $X$ and $Z$ are binary, one can easily see that every non-deterministic relation between $X$ and $Z$ destroys pairwise purity.

The following result shows that pairwise purity can still hold if only some of the causal paths contain a low range variable: (see statement (iii) in Hypothesis 1):

**Lemma 9 (purity after marginalization)**
Let $X := (X_1, X_2, \ldots, X_n)$ where the range of each $X_j$ is a finite subset of $\mathbb{R}$. Let $P(Y|X)$ be given by the linear additive noise model

$$Y = \sum_j w_j X_j + E \text{ with } E \perp\!\!\!\perp X,$$

with $w_1 \neq 0$ and $E$ satisfies the asymptotical condition (3). If $P(X_2, X_3, \ldots, X_n | X_1)$ is strictly positive, then $P(Y|X_1)$ is also pairwise pure.

Proof: We have $P(Y|X_1) = \sum_{x_{2..n}} P(Y|X_1, x_{2..n}) \times P(x_{2..n}|X_1)$, where $x_{2..n} := (x_2, \ldots, x_n)$. Assume $w_1 > 0$ and let $x_{2...n}^{\max}$ be the $n-1$ tuple that maximizes $\sum_{j=2}^{n} w_j x_j$. If there are more than one options, we can choose one of them. Likewise, let $x_{2...n}^{\min}$ minimize the expression. The quotient $p(y|x_1)/p(y|x_1')$ is for $y \to \infty$ asymptotically dominated by $p(y|x_1, x_{2...n}^{\max})/p(y|x_1', x_{2...n}^{\max})$, which satisfies (3) by assumption. Likewise, $p(y|x_1, x_{2...n}^{\min})/p(y|x_1', x_{2...n}^{\min})$, dominates the asymptotics for $y \to -\infty$. For fixed $x_{2..n}^{max}$ and fixed $x_{2..n}^{min}$, the density $p(y|x_1, x_{2...n}^{\max})$ and

$p(y|x_1, x_{2...n}^{min})$ define pairwise pure additive noise models. Then the asymptotics $y \to \pm\infty$ shows pairwise purity. Assuming $w_1 < 0$ only swaps the asymptotics for $y \to \infty$ and $y \to -\infty$. □

## 3 Evidence against "direct" causal relevance: dimension estimation

While the concept of purity deals with the location of the distributions $P(Y|x)$ relative to the boundaries of the simplex $\mathcal{P}_Y$, we now explore linear dependence between the different $P(Y|x)$. The idea is that linear dependence is non-generic and thus require an explanation in terms of a causal mechanism that implies low dimensionality. A simple kind of dependence is given by the case where $P(Y|x) = P(Y|x')$ for some $x \neq x'$. This occurs, for instance, if $X$ is a vector-valued variable where only some of the components influence $Y$. One could also think of a thresholding mechanism where $P(Y|x) = Q_0$ for all $x \leq x_0$ and $P(Y|x) = Q_1$ for all $x > x_0$. The following vague statement will be the leading idea of this section:

**Hypothesis 2 (linear independence)**
*Let $N(P(Y|X))$ be the number of different distributions $P(Y|x)$. Generically, the distributions $\{P(Y|x)\}_{x \in \mathcal{X}}$ span a vector space of dimension $N(P(Y|X))$ whereas linear dependence requires a causal explanation.*

We will now explore possible explanations, without claiming that our list is complete.

**Lemma 10 (finite rank of concatenations)**
*Let $Z$ be a variable with $X \perp\!\!\!\perp Y | Z$. Let $D(P(Y|X))$ denote the dimension of the span of $\{P(Y|x)\}_{x \in \mathcal{X}}$. Then*

$$D(P(Y|X)) \leq \min\{D(P(Y|Z)), D(P(Z|X))\}.$$

*Let $D_c(P(Y|X))$ denote the smallest number of points in $\mathcal{P}$ containing all $P(Y|x)$ in their convex span. Then,*

$$D_c(P(Y|X)) \leq \min\{D_c(P(Y|Z)), D_c(P(Z|X))\}.$$

To see that $D$ and $D_c$ need not coincide we consider the tetrahedron, which visualizes the set $\mathcal{P}_\mathcal{Y}$ for $|\mathcal{Y}| = 4$. One can cut the tetrahedron such that the cut is a square. Let $P(Y|x_j)$ for $j = 1, \ldots, 4$ be the 4 corners of this square. Although they span a 3-dimensional space, the tetrahedron does not contain a triple of points whose convex hull contains all $P(Y|x_j)$.

Proof: (of Lemma 10) Due to $P(Y|x) = \int P(Y|z) dP(z|x)$, every $P(Y|x)$ is in the convex hull of $\{P(Y|z)\}_{z \in \mathcal{Z}}$. This shows $D(P(Y|X)) \leq D(P(Y|Z))$ and $D_c(P(Y|X)) \leq D_c(P(Y|Z))$.

If every $P(Z|x)$ is in the linear span of $(Q_j)_{j=1}^d \subset \mathcal{P}_Z$ we can write $P(Z|x) = \sum_{j=1}^d \alpha_j(x) Q_j$ for some $\alpha_j(x) \in \mathbb{R}$ with $\sum_j \alpha_j(x) = 1$. Hence, $P(Y|x) = \sum_{j=1}^d \alpha_j(x) \int P(Y|z) dQ_j(z)$. This shows $D(P(Y|X)) \leq D(P(Z|X))$. If all $P(Z|x)$ are in the convex hull of all $Q_j$ then we can choose all $\alpha_j(x)$ to be non-negative, hence $D_c(P(Y|X)) \leq D_c(P(Z|X))$. □

The simplest explanation for $D_c(P(Y|X))$ having lower rank than $N(P(Y|X))$ is that the dependences between $X$ and $Y$ are due to a variable $Z$ with range smaller than $N(P(Y|X))$:

**Corollary 2 (finite range $Z$)**
*If $X \perp\!\!\!\perp Y | Z$ then $D_c(P(Y|X)) \leq |\mathcal{Z}|$.*

However, $Z$ need not have finite range. We could also think of $P(Y|Z)$ being a mechanism where some $P(Y|z)$ coincide, e.g., by a thresholding mechanism. The results above suggest to consider $D(P(Y|X))$ being smaller than $N(P(Y|X))$ as a strong hint against a "direct" causal connection.

To explore the mathematical relation between purity and dimension $D$ we mention the following result although its implication for *pairwise* purity is trivial:

**Lemma 11 (upper bound on purity)**
*If $P(Y|X)$ is $k$-wise pure then the span of all $P(Y|x)$ is at least $k$-dimensional.*

Proof: If $P(Y|x_1), \ldots, P(Y|x_k)$ are linearly dependent, we can find $\alpha_j \in \mathbb{R}$ with $\sum_j \alpha_j = 0$ such that $\sum_{j=1}^k \alpha_j P(Y|x_j) = 0$. At least one of these coefficients must be negative, assume wlog $\alpha_1 < 0$. Then, $P(Y|x_2) + \sum_{j=1}^k \alpha_j P(Y|x_j) = P(Y|x_2) \in \mathcal{P}_\mathcal{Y}$ contains at least one negative coefficient, namely $\alpha_1$, therefore $P(Y|X)$ is not $k$-wise pure. □

Intuitively, one expects pairwise purity of $P(Y|X)$ to be less likely if $D(P(Y|X))$ is small. The mathematical bounds are, however, weak: Corollary 1 shows that the minimal size of $\mathcal{Y}$ required for pairwise purity grows slowly in $|\mathcal{X}|$ because the combinatorial number $m(|\mathcal{Y}|)$ grows fast.

The following theorem connects the concepts of purity and dimension $D$. It shows that we can get more information about a latent binary cause than just knowing that it exists.

**Theorem 3 (reconstruct latent binary cause)**
*Let $Z$ be a binary variable such that $X \perp\!\!\!\perp Y | Z$ and $P(Y|Z)$ is pairwise pure. Then $P(X, Z, Y)$ is determined by $P(X, Y)$ up to inverting $Z$.*

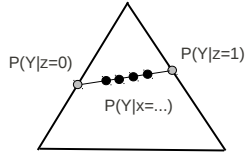

Figure 2: Reconstruction of $P(X,Y,Z)$: We first obtain $P(Y|z)$ (grey circles) as the points where the line connecting all $P(Y|x)$ crosses the boundary of the simplex. Then the location of each $P(Y|x)$ (black circles) determines each $P(Z|x)$.

Proof: Given any two distinct points $P(Y|x_1)$ and $P(Y|x_2)$, all other $P(Y|x)$ and $P(Y|z)$ lie on the line $\mu P(Y|x_1) + (1-\mu)P(Y|x_2)$ with $\mu \in \mathbb{R}$ (see Fig. 2). Let $\mu_0$ and $\mu_1$ be the supremum and infimum, respectively, of all $\mu$ for which $\mu P(Y|x_1) + (1-\mu)P(Y|x_2) \in \mathcal{P}_Y$. Then, pairwise purity implies that $P(Y|z) = \mu_z P(Y|x_1) + (1-\mu_z)P(Y|x_2)$ for $z = 0, 1$, up to inverting $Z$. $P(Z|X)$ is uniquely determined by $P(Y|x) = P(Y|z=0)p(z=0|x) + P(Y|z=1)p(z=1|x)$, and we have $P(X,Z,Y) = P(X)P(Z|X)P(Y|Z)$. $\square$

## 4 Purity test and dimension estimation from finite data

**Testing pairwise purity**
We propose two alternative approaches to test purity from finite data. In our experiments we only considered the second approach that is based on density estimation, as this variant turned out to be more reliable and applicable also to small sample sizes.

(1) To show that $P(Y|X)$ is close to being pairwise pure we only need to show that $P(Y \in S|x)/P(Y \in S|x')$ is small for some subset $S \subset \mathcal{Y}$, for all pairs $(x,x')$ (cf Lemma 1). If the set $S$ has been chosen before seeing the data, we can also conclude (with high probability after large sampling) that this quotient is small after observing that the quotient of relative frequencies $\hat{P}(Y \in S|x)/\hat{P}(Y \in S|x')$ is small. Moreover, VC-learning theory [Vapnik, 1998] shows that the same conclusion is allowed when we find one set $S_j \in (S_j)_{j \in J}$ having small quotient provided that $(S_j)_{j \in J}$ is a family of sets whose VC-dimension is sufficiently low compared to the sample size. We have derived a precise version of this statement, but omitted it due to space constraints and because the bounds require large sample sizes that exceed values that are reached in practice.

(2) Note that, in order to reject purity from finite data, smoothness assumptions on the density are needed, because small regions of density zero would be hard to detect. Hence, to derive a practical test of purity for finitely many samples, we propose to employ kernel density estimation (in our experiments we use a Gaussian kernel) to estimate the density of the observed $Y$ conditioned on the state of $X$, $\hat{P}(Y|x)$. To decide whether we consider $P(Y|X)$ to be pairwise pure, we then need to define a region over which we minimize the ratios of $\hat{p}(y|x)/\hat{p}(y|x')$, for all pairs $(x,x')$. Minimizing over all possible $y$ is not feasible because the density estimate is unreliable in areas far from observed data points. Hence, we reverted to a pragmatic solution, constraining $y \in \Psi$, with $\Psi$ being a set of equally spaced points in the interval $[y_{\min}, y_{\max}]$, where $y_{\min}$ and $y_{\max}$ denote the minimum and the maximum of all observed $y$-values. This approach also justifies our choice of a Gaussian kernel, despite the dependence of purity on the tails of the conditional distribution (see Lemma 4 and 7). To decide whether we consider $P(Y|X)$ to be pairwise pure, we need to compute $\min_{y \in \Psi}(\hat{p}(y|x)/\hat{p}(y|x'))$ for all pairs $(x,x')$. We will refer to the $\max_{\{(x,x')\}}(\min_{y \in \Psi}(\hat{p}(y|x)/\hat{p}(y|x')))$ as the purity ratio of the conditional $P(Y|X)$.

**Estimating dimension**
To estimate the dimension of the space spanned by all $P(Y|x)$, we propose to represent each $P(Y|x)$ as a vector in the same Hilbert space $\mathcal{H}$. Then the dimension is given by the rank of the Gram matrix

$$M := (\langle P(Y|x), P(Y|x') \rangle)_{x, x' \in \mathcal{X}}.$$

Smola et al. [2007] proposed a method for representing distributions as vectors in reproducing kernel Hilbert spaces (RKHS). We will describe this method because it allows for simple estimators $\hat{P}(Y|x)$ which converge to $P(Y|x)$ with respect to the Hilbert space norm. A positive definite kernel $k : \mathcal{Y} \times \mathcal{Y} \to \mathbb{R}$ defines a Hilbert space of functions given by the completion of the linear span of the functions $y \mapsto k(y, y')$ with $y' \in \mathcal{Y}$ ($k(., y')$ for short). The inner product is defined [Schölkopf and Smola, 2002] via $\langle k(., y), k(., y') \rangle = k(y, y')$. Then every distribution $P(Y|x)$ defines a vector via $P(Y|x) := \int k(., y) dP(y|x)$. For every $x$, we are given the observations $y_1^x, \ldots, y_{n_x}^x$ and have the estimator

$$\hat{P}(Y|x) := \frac{1}{n_x} \sum_{j=1}^{n_x} k(., y_j^x).$$

This defines the estimated Gram matrix $\hat{M}$ with entries

$$\hat{M}_{x,x'} := \frac{1}{n_x n_{x'}} \sum_{i=1}^{n_x} \sum_{j=1}^{n_{x'}} k(y_i^x, y_j^{x'}).$$

To estimate the rank of $M$ we have to decide whether the small eigenvalues of $\hat{M}$ are due to statistical estimations from deriving bounds on $\|\hat{P}(Y|x) - P(Y|x)\|$.

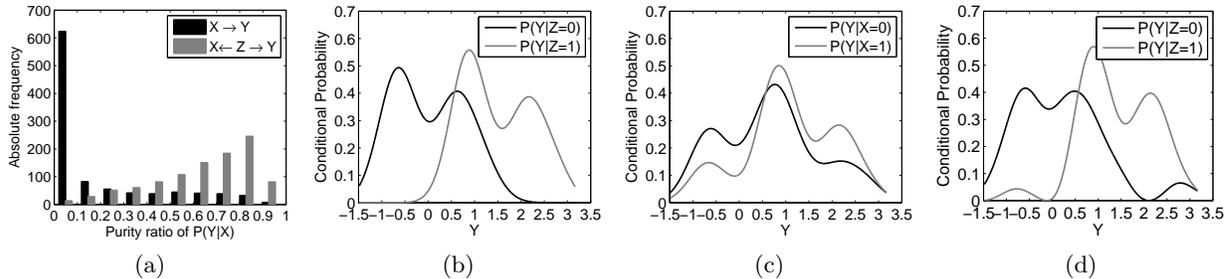

(a) (b) (c) (d)

Figure 3: (a) Histogram of the purity ratios for the two experimental settings; for the simulation setting $X \to Y$ the purity ratio values are closer to zero than for $X \leftarrow Z \to Y$. (b) simulated $P(Y|Z)$ (unobserved), (c) $P(Y|X)$ (observed) and (d) reconstructed $P(Y|Z)$, respectively, for the simulation setting $X \leftarrow Z \to Y$. The quality of the reconstructed conditional can be evaluated by comparing (d) to the true simulated conditional in (b).

## 5 Experiments

### 5.1 Synthetic examples

**Testing pairwise purity**
To test for pairwise pure conditionals, we considered variables $X, Y, Z$ with the following ranges: $\mathcal{X} := \{0, 1\}$, $\mathcal{Z} := \{0, 1\}$, $\mathcal{Y} := \mathbb{R}$. In the first experiment we simulated the setting $X \to Y$. For that, we considered $P(Y|X)$ to be a linear additive noise model $Y = wX + E$, where $w$ is a weight drawn from a zero mean Gaussian with unit variance. To demonstrate that pure conditionals are feasible in settings where $P(Y|x)$ is multi-modal (see note after Lemma 5), we chose $E$ to be distributed according to a mixture of two Gaussians. The model thus satisfies the assumptions of Lemma 4. In each simulation run, we drew 1000 independent samples, first drawing $x$ from a Bernoulli distribution with success probability chosen uniformly at random from $[0, 1]$, and then sampling $y$ from the conditional distribution $P(Y|x)$.

In the second experiment, we considered $X \leftarrow Z \to Y$, first simulating the unobserved latent variable $Z$, again choosing $P(Z = 1)$ at random as before for $P(X)$. Then, we used a linear additive noise model $Y = wZ + E$ (as above) to draw $y$ from $P(Y|z)$. The observed variable $X$ was simulated using randomly chosen transition probabilities $P(X|Z)$. Specifically, $P(X = 0|Z = 0)$ and $P(X = 0|Z = 1)$ where drawn uniformly from $(0, 1)$.

To test for purity, we computed the purity ratio of $P(Y|X)$ (see Sec. 4) for 1000 simulation runs of each of the two experiments described above. Figure 3(a) depicts a histogram of the estimated ratios for both settings. It is noticeable that purity ratios of $P(Y|X)$ from the first experiment ($Y$ was generated from $X$) are predominantly smaller than 0.1, whereas ratios from the second experiment ($Y$ was generated from $Z$) tend to yield higher values.

**Reconstructing unobserved binary causes**
In the proof of Theorem 3, we showed how to reconstruct $P(Y|Z)$, where $Z$ is a latent binary cause with $P(Y|Z)$ pairwise pure and $X \perp\!\!\!\perp Y | Z$. Here, we applied this approach to the synthetic data from a single run of the second experiment (setting the parameters accordingly) in order to illustrate the reconstruction procedure by an example. Figures 3(b), 3(c), 3(d) show the true simulated conditional $P(Y|Z)$ (unobserved), the observed $P(Y|X)$ and the reconstructed conditional $P(Y|Z)$, respectively. Comparing the reconstructed conditional (Fig. 3(d)) with the true simulated one (Fig. 3(b)), we observe an encouraging reconstruction quality.

### 5.2 Applications to statistical genetics

Next, we applied our method to a problem in statistical genetics, as already mentioned in the motivation of this work. Reliable ground truth is difficult to obtain in genetic studies, and hence, following previous work (e.g. [Platt et al., 2010]), we considered realistic simulated settings. Our simulation was based on data form a 250K SNP chip from *Arabidopsis*, consisting of 1200 samples (downloaded http://walnut.usc.edu/2010/data/250k-data-version-3.06). Hence, only the dependence between real genetic data and phenotype measurements was simulated, whereas the joint distribution of SNPs was based on real data.

**Identifying causal SNPs using purity and correlation**
We investigated to what extend the purity ratio is indicative of a causal relationship between a SNP and a phenotype. As a comparison, we also considered correlation, a basic measure of association that is commonly used in genetical studies [Balding, 2007].

We again considered two experimental settings. First, we simulated SNP-phenotype associations according to the setting $X \to Y$, first choosing a SNP $X$ at random from the 250K SNPs, and then generating the phenotype $Y$ from a linear additive noise model $Y = wX + E$ as before (see Section 5.1), where $E$ here follows a Gaussian distribution.

Analogously, we simulated associations according to the setting $X \leftrightarrow Z \to Y$. Here, $Z$ is a SNP randomly selected among the set of all SNPs. We generated the phenotype measurement $y$ from $P(Y|z)$, again employing the same linear additive noise model, $Y = wZ + E$. The non-causal SNP $X$ was chosen to be next to $Z$. This choice is motivated by the strong correlation between nearby SNPs, leading to an ambiguity as to which SNP is the truly causal one among a set of SNPs that may all exhibit strong correlation to the same phenotype.

In total, we generated 1000 SNP/phenotype pairs according to each of the two settings described above. For each pair we estimated the purity ratio of $P(Y|X)$ as well as the correlation coefficient $r^2(X,Y)$. Fig. 4 shows the relationship between the correlation coefficients and the negative logarithm of the corresponding purity ratios, for both experimental settings. Note that, high correlation coefficients were observed in both settings, while purity appeared to discriminate between the settings. Even for strongly correlated non-causal SNPs, the purity (the negative logarithm of the purity ratio) remained low and hence did not give false evidence for a causal link.

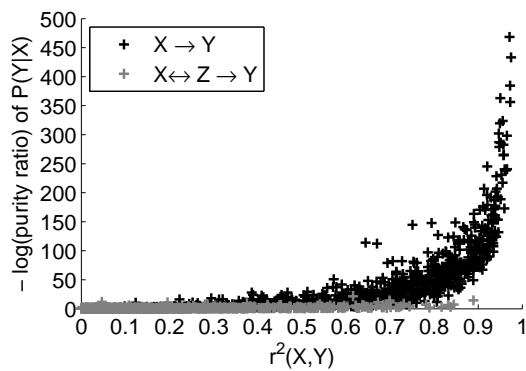

Figure 4: Scatter plot of the correlation between each SNP and its phenotype versus the negative logarithm of the purity ratio of $P(Y|X)$. Shown are SNPs that are causal (black) and non-causal (grey) separately.

**High correlation between the phenotype and the non-causal SNPs**

Misleading correlation structure is a challenge in real association studies. A recent study in [Platt et al., 2010] investigates very similar simulated models to highlight the risk of positively misleading answers from correlation analyses. We designed this experiment such that the correlation between a non-causal SNP and its corresponding simulated phenotype can be higher than the correlation between the causal SNP and the phenotype.

We first simulated causal pairs $(X, Y)$ (SNP, phenotype), generating $Y$ as

$$Y = w_1 X + w_2 V + E,$$

where $X$ is any random SNP, $V$ is simulated as a corrupted version of another SNP located far from $X$ and $w_2 = 2w_1$. Accordingly, we generated non-causal $X$, $Y$ (SNP, phenotype), first choosing $X$ randomly from the set of all SNPs and then generating $Y$ as

$$Y = w_1 Z + w_2 V + E,$$

where $Z$ is simulated as a corrupted version of $X$, $V$ is a SNP located far from $X$ and $w_1 = 2w_2$. To simulate a corrupted version of a SNP, we inverted a certain percentage (corruption level) of its samples (here, $Z := X \oplus C$, with $\mathcal{C} := \{0,1\}$ and $P(C = 1)$ being the corruption level).

Using the above setting for the weights of the models, we often got high correlations between simulated non-causal SNPs and their corresponding phenotypes and low correlations between simulated causal SNPs and their corresponding phenotype, which can be misleading for the inference of the causal direction. We compared the ability of purity and correlation to classify a SNP as causal or non-causal after generating 1000 causal SNP/phenotype and 1000 non-causal SNP/phenotype pairs. Fig. 5 shows the area under the receiver operating characteristic (ROC) curve (AUC) for both methods (purity and correlation) and for a range of different corruption levels. We can observe that purity consistently makes more accurate decisions than naive correlation analysis. In particular, for the limit of zero corruption both methods failed due to the strong coupling of non-causal SNPs with a simulated cause of their corresponding $Y$. As stated in the Introduction, it is impossible to distinguish between too strongly coupled variables. In the regime of higher corruption, purity outperformed the correlation-based approach. Finally, in the limit of maximal corruption, both methods performed equally well, since the non-causal SNPs were not correlated anymore to the phenotype.

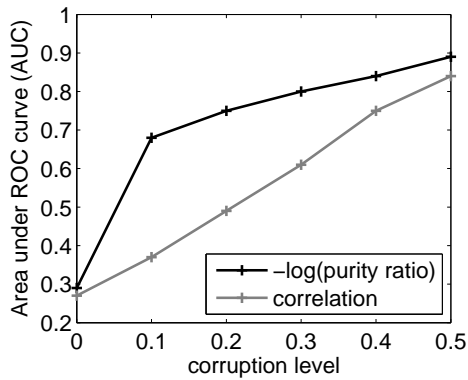

Figure 5: Area under the receiver operating characteristic (AUC) as a function of the corruption level, both using purity and correlation correlation to identify a causal SNP. In the deterministic case (corruption=0) both methods fail, as non-causal SNPs are deterministically coupled wit a simulated cause of $Y$. In the regime of high corruption levels (0.5), both methods perform equally well, since non-causal SNPs are not correlated anymore with the phenotype. In the relevant regime of an intermediate level of corruption, purity clearly outperforms the correlation measure.

## 6  Discussion and Conclusion

In this paper, we proposed a novel approach to collect evidence for or against a "direct" causal link $X \to Y$ between two statistically dependent observed variables. In case of latent causes with low range, the method builds on an informative property of the conditional densities $P(Y|X)$, which we call purity. The characterization of a conditional as pure is based on the location of the different $P(Y|x)$ in the simplex of probability distributions of $Y$. We showed that many common causal relationships (e.g., a significant subset of additive noise models) lead to pairwise pure conditionals, and that confounding factors are likely to lead to non-pure conditionals if a low range variable is contained in all unblocked paths between $X$ and $Y$. We employed these theoretical results to construct an empirical test for causal relationships, based on kernel density estimation. We conducted experiments to estimate purity from finite data, and to reconstruct the conditional distribution $P(Y|Z)$, with $Z$ being an unobserved binary cause, from the observed conditionals $P(Y|X)$.

There is a broad range of potential applications suitable for our approach. Here, we considered a problem from statistical genetics that served as a motivation for our method. In this setting, our goal was to identify those genetic markers that are causally relevant for a certain phenotype. In the experiments we showed that the proposed approach based on purity can distinguish cause-effect relations from spurious links with higher accuracy than a standard correlation measure.

We believe that our approach provides the foundation for an interesting direction of research. Several important questions remain to be addressed. For example, efficient estimators of purity should be developed in future work.

### Acknowledgements

DJ has been supported by the DFG (SPP 1395).

## References

D. Balding. *Handbook of statistical genetics*. Wiley-Interscience, 2007.

P. Hoyer, D. Janzing, J. Mooij, J. Peters, and B. Schölkopf. Nonlinear causal discovery with additive noise models. In *Proceedings NIPS 2008*. MIT Press, 2009.

D. Janzing, J. Peters, J. Mooij, and B. Schölkopf. Identifying latent confounders using additive noise models. In *Proceedings UAI 2009*, pages 249–257.

J. Pearl. *Causality*. Cambridge University Press, 2000.

A. Platt, B. Vilhjalmsson, and M. Nordborg. Conditions under which genome-wide association studies will be positively misleading. *Genetics*, 186(3):1045, 2010.

H. Reichenbach. *The direction of time*. University of California Press, Berkeley, 1956.

B. Schölkopf and A. Smola. *Learning with kernels*. MIT Press, Cambridge, MA, 2002.

S. Shimizu, P. Hoyer, and A. Hyvärinen. Estimation of linear non-gaussian acyclic models for latent factors. *Neurocomputing*, 72(7–9):2024–2027, 2009.

A. Smola, A. Gretton, L. Song, and B. Schölkopf. A Hilbert space embedding for distributions . *Lecture Notes in Computer Science*, 4755:40–41, 2007.

V. Vapnik. *Statistical learning theory*. John Wileys & Sons, New York, 1998.

K. Zhang and A. Hyvärinen. On the identifiability of the post-nonlinear causal model. In *Proceedings UAI 2009*, pages 647–655.